\documentclass[letterpaper, 10 pt, conference]{ieeeconf}
\IEEEoverridecommandlockouts
\usepackage{amsmath,amssymb,graphicx,booktabs,url,xcolor}
\usepackage{newtxtext,newtxmath}
\usepackage[letterpaper,left=0.65in,right=0.65in,top=0.65in,bottom=0.7in,footskip=0.3in]{geometry}
\usepackage[hidelinks]{hyperref}
\hypersetup{pdftitle={The Unexpired Plan: A Free Monitor for Accelerated Diffusion Policies},
pdfauthor={Yi Zhao and Sebastian Scherer}}

\AtBeginDocument{\setlength{\abovedisplayskip}{6.5pt plus 2pt minus 2pt}%
\setlength{\belowdisplayskip}{6.5pt plus 2pt minus 2pt}%
\setlength{\abovedisplayshortskip}{3pt plus 2pt}%
\setlength{\belowdisplayshortskip}{4.5pt plus 2pt minus 2pt}}
\newcommand{\pp}{\,\mathrm{pp}}
\AtBeginDocument{\let\rawref\ref\renewcommand{\ref}[1]{\mbox{\rawref{#1}}}}
\newcommand{\best}[1]{\textbf{\boldmath#1}}
\newcommand{\second}[1]{\underline{#1}}
\newcommand{\grey}[1]{\textcolor[rgb]{0.45,0.45,0.45}{#1}}
\usepackage{stfloats}
\usepackage[nospread]{flushend}
\usepackage[all]{hypcap}
\makeatletter
\renewcommand{\@IEEEauthorblockNstyle}{\normalfont\fontsize{14}{17}\selectfont}
\renewcommand{\@IEEEauthorblockAstyle}{\normalfont\fontsize{11}{14}\selectfont}
\renewcommand{\@IEEEauthorblockAtopspace}{0.5ex}
\let\paper@makecaption\@makecaption
\long\def\@makecaption#1#2{{\let\footnotesize\small\paper@makecaption{#1}{#2}}}
\makeatother
\newcommand{\papertable}{\renewcommand{\arraystretch}{1.16}}

\title{The Unexpired Plan: A Free Monitor\\
for Accelerated Diffusion Policies}

\author{\authorblockN{%
\begin{tabular}{@{}c@{\hspace{3em}}c@{}}
Yi Zhao & Sebastian Scherer\\[-1pt]
{\fontsize{10.5}{13}\selectfont\href{mailto:yizhao5@andrew.cmu.edu}{yizhao5@andrew.cmu.edu}} &
{\fontsize{10.5}{13}\selectfont\href{mailto:basti@andrew.cmu.edu}{basti@andrew.cmu.edu}}
\end{tabular}}
\authorblockA{AirLab, The Robotics Institute\\
Carnegie Mellon University, Pittsburgh, PA, USA\\[3pt]
\href{https://yizhaojasper.com/unexpired-plan/}{\textbf{Project website:} \texttt{yizhaojasper.com/unexpired-plan/}}}}

\begin{document}
\maketitle
\thispagestyle{plain}

\begin{abstract}
Training-free acceleration of a diffusion policy is accepted when an internal similarity signal
reports that the shortcut changed nothing. We price every monitor a control loop can afford in
closed-loop success rather than feature distance, over $114$ accelerator configurations and four
policy families: two accelerators each clear their own gate's bar and log every reuse as certified,
while on one task one finishes every episode and the other none. A monitor is two
designs, not one --- the statistic it reads, and what it does when that statistic fires.
Published gates re-arm after every rejection, and under that response even an oracle handed
every forward pass and the exact local action error loses twenty points; absorbing
the same statistic costs one point, and most of its speed. What makes a response that never
forgets affordable is a statistic that rarely fires, and what it must measure is deviation from
the policy the accelerator replaced --- which a chunked policy has already paid for, its last
plan not yet expired and free to read. Guarding every call this way cuts per-call compute by
$1.55$--$3.09\times$, where any reference-requiring check at the same coverage would have to
stop accelerating altogether. On three of our four families the schedule alone already holds
the pre-stated $\pm2$-point margin. What the monitor is measurably worth shows in three places:
on the fourth family, where it rescues the candidate selection landed on; on four configurations
it did not select; and on a contact-rich fifth family, chosen where the schedule was expected to
fail and run after every design choice was frozen, where no unmonitored arm at its speed holds
the margin and the monitored one does.
\end{abstract}

\begin{figure}[!b]\centering
\includegraphics[width=\columnwidth]{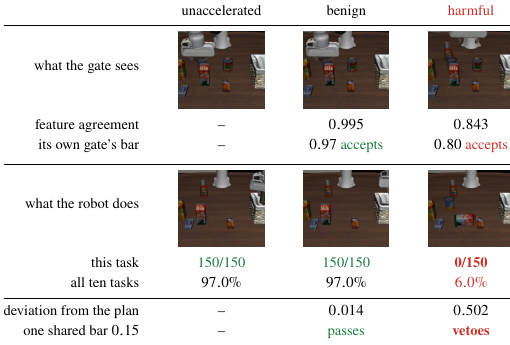}
\caption{\textbf{A gate cannot be read from its own log.} One LIBERO-10 task, same initial
states: each accelerator scores its own cached quantity against its own bar, so both certify
every reuse, while one finishes every episode and the other none. The last rows read the
quantity \S\ref{sec:method} monitors against one bar shared by all.}
\label{fig:teaser}
\end{figure}

\section{Introduction}
Diffusion policies spend most of their inference budget recomputing what changed little since
the previous control call; training-free accelerators exploit that by caching features,
skipping denoising steps or reusing action chunks.

Knowing when a shortcut is admissible is the unsolved half, and the field's answer is to let
the accelerator watch itself: compare a cached quantity against a freshly computed one and
reuse while they agree. Closed-loop success is reported for the setting that ships, but the rule deciding
each reuse is argued in feature space, a weak predictor of closed loop~\cite{clgap}. Validated
in closed loop instead, the outcome is not a matter of degree: a cross-call cache in our sweep
drives LIBERO-10 success from $97\%$ to $6\%$ (Fig.~\ref{fig:teaser}) while certifying every
one of the $161{,}000$ reuses it \emph{logs} across all its episodes.

Rather than report one failure we measured the whole design space. Section~\ref{sec:map} prices
every monitor a control loop can afford, over $114$ configurations from four policy families,
and the map is strongly negative: at a matched acceptance rate an exogenous gate does not beat
discarding forwards at random, and an oracle handed every forward pass and the exact deviation still
loses twenty points. What failed there is the \emph{response}, not the signal, and the
gating literature has been benchmarking signals. On the signal axis every free monitor Table~\ref{tab:map} prices fails too: the
one statistic that separates catastrophic from benign configurations is the deviation from
the policy the accelerator replaced, exactly the quantity a free monitor cannot hold.

It can be held, if the loop is read carefully. A chunked policy predicts $H$ actions and
executes only $H_{\mathrm{exec}}<H$ of them, so a reference evaluation leaves a plan still valid
for the next $\lfloor H/H_{\mathrm{exec}}\rfloor-1$ calls. Comparing the accelerator against
that unexpired plan costs no forward pass and carries real reference provenance; at the one call in $m$
with no plan left we run the reference and execute it.

\textbf{Contributions.} A measured signal $\times$ response map, priced against a matched-rate
coin and a local-deviation oracle at full compute (\S\ref{sec:map}); a free monitor that reads the replaced policy,
with an exact coverage law (\S\ref{sec:free}), priced on five families (\S\ref{sec:exp}); and a
coverage--speed--risk frontier on
which full coverage is a corner, at a ceiling of $m$ --- a policy that plans two calls ahead
is never guarded past $2\times$ (\S\ref{sec:cost}, \S\ref{sec:frontier}).
\section{Related Work}
\textbf{Training-free acceleration.} Feature caching, step skipping and attention reuse
accelerate video diffusion transformers per layer or per
step~\cite{deepcache,worldcache_eccv,magcache,teacache}. On robot policies, accelerators cache
visual tokens~\cite{gatedvlacache}, retrieve and refine chunks~\cite{actioncache}, carry a cache
across inference chunks~\cite{c3ache}, re-emit the previous chunk when consecutive chunks
agree~\cite{flashvla}, or exit early over depth~\cite{deervla}. These are the population we
measure, not rivals we out-cache. Distillation reaches one step but retrains the
weights~\cite{consistencypolicy}, and a learned cache schedule trains a router~\cite{l2c};
both are outside our setting.

\textbf{Runtime monitoring.} The closest work shows that gate provenance governs the
closed-loop reliability of training-free skipping~\cite{gatenotcache}; its actuation-slack
refresh spends a forward to repair the skip, where we execute that forward and harvest its
plan. Action-JND learns token tolerances from a frozen policy~\cite{actionjnd}; ours needs no
learned estimator. Denoising-variance prefixes~\cite{dvac} and observation-conditioned
caches~\cite{adacache} are the family we sweep in \S\ref{sec:map}. The overlap of successive plans is
itself well used: temporal ensembling averages what successive chunks predict for one
index~\cite{act}, bidirectional decoding scores their agreement to choose a sample~\cite{bid},
STAC~\cite{stac} turns it into a runtime failure alarm, and chunked controllers read it to time
a replan~\cite{flashvla}. We neither average nor sample from it; we compare against it and veto,
and what is new is what that costs: STAC draws fresh samples at both calls, where the unexpired
plan is a reference evaluation already paid for. Table~\ref{tab:main} prices the difference.
Speculative decoding~\cite{specdec} verifies each draft token \emph{exactly} against the
target it replaces; ours is neither exact nor contemporaneous, which is why
\S\ref{sec:frontier} prices per-call misses (Eq.~\eqref{eq:feas}).

\textbf{Offline search.} Rollout-success search costs a GPU-day per task~\cite{evo}; we score
deviations, never success (\S\ref{sec:sel}). Real-time chunking alters chunk
timing~\cite{rtc}, outside our fixed schedule.

\section{Measuring what a monitor is worth}\label{sec:map}
\textbf{Protocol.} Every configuration is rolled out in closed loop whatever its monitor
predicted, paired against the unaccelerated policy on identical initial states: $114$
configurations over a video--action diffusion policy on LIBERO~\cite{cosmospolicy,libero},
$\pi_{0.5}$ on LIBERO-10~\cite{pi05}, a Diffusion Policy on PushT~\cite{diffusionpolicy}, and
3D Diffusion Policy on MetaWorld~\cite{dp3,metaworld}. Monitors are instrumented on every
evaluation, not only when they accept. Two responses are crossed throughout: \emph{re-arming}
restores permission after each rejection, \emph{absorbing} does not. Rows of
Table~\ref{tab:map} that share a \emph{base} guard the same accelerator on the same family, so
the coin, the observation gate and the oracle are matched to each other and the last block
guards the first row's cache in place of its own gate; every comparison across rows below is within one such group.
Under \emph{absorbing} a veto retires the accelerator, or on the ladder row a rung; the last
block's accelerated share is capped at $(m-1)/m$ by its $m$-periodic reference; and a coin
matched to a gate's \emph{retirement} is absorbing by construction, hence two empty cells.

\textbf{A gate cannot be read from its own log.} A gate records a value only when it accepts,
so its maximum is bounded by its own threshold: the cross-call cache's gate certifies every
reuse it logs and, once instrumented on every call, accepts $53.8\%$ of them.

\begin{table*}[t]\centering\papertable\small
\caption{Every monitor a control loop can afford, priced in closed-loop success. $\Delta$SR is
paired against \emph{base} on identical initial states over the episode count of \S\ref{sec:exp}'s protocol; the half-widths $h=\pm1.1\pp$
at base $98.2$ and $h=\pm0.9\pp$ at $97.0$ hold near base only.
\emph{accel.} is the share that ran accelerated under \emph{re-arming}; \best{bold} marks our
rows.}
\label{tab:map}
\setlength{\tabcolsep}{5pt}
\begin{tabular}{@{}lllcccc@{}}
\toprule
& & & & & \multicolumn{2}{c}{$\Delta$SR by response}\\
\cmidrule(lr){6-7}
What the statistic measures & Monitor & Permission & accel.\ (\%) & base & re-arming & absorbing\\
\midrule
\multicolumn{7}{@{}l}{\emph{Free, but not the deviation the accelerator introduced}}\\
agreement of cached features & cross-call gate, deployed & admits & $53.8\%$ & $97.0$ & $-91.0$ & $-88.1$\\
agreement of cached features & same statistic, veto-only & vetoes only & $61.4\%$ & $97.0$ & $-92.3$ & $-89.4$\\
the call's own inputs & observation gate, safe settings & admits & $1.3\%$ & $98.2$ & $-6.4$ & $-5.9$\\
the call's own inputs & observation gate, at the coin's rate & admits & $46.2\%$ & $98.2$ & $-91.1$ & $-88.4$\\
the call's own inputs & observation gate, useful settings & admits & $78.5\%$ & $98.2$ & $-97.6$ & $-87.6$\\
\emph{nothing} & \emph{random reuse, matched rate} & \emph{admits} & $46.2\%$ & $98.2$ & $-88.6$ & $-0.7$\\
\emph{nothing} & \emph{random, matched retirement} & \emph{admits} & --- & $98.2$ & --- & $-87.4$\\
the policy's own value estimate & value head & admits & $75.9\%$ & $97.0$ & $-20.6$ & $-18.9$\\
\midrule
\multicolumn{7}{@{}l}{\emph{The deviation itself, but only by recomputing what the accelerator skipped}}\\
exact local action error & local-deviation oracle, $\tau{=}0.2$ & admits & $15.3\%$ & $98.2$ & $-20.6$ & $-1.1$\\
\midrule
\multicolumn{7}{@{}l}{\emph{The deviation, free, read off the replaced policy itself --- on the first row's cache ($\pi_{0.5}$, $m{=}5$)}}\\
\best{its last plan, not yet expired} & \best{unexpired plan alone (ours)} & \best{vetoes only} & $33.7\%$ & $97.0$ & $-0.4$ & $-0.2$\\
\best{a repeat in the emitted chunk} & \best{chunk repetition (ours)} & \best{vetoes only} & $80.0\%$ & $97.0$ & $-91.6$ & $-91.6$\\
\best{both, on a ladder (\S\ref{sec:sel})} & \best{what we deploy} & \best{vetoes only} & $33.7\%$ & $97.0$ & \best{$-0.4$} & \best{$-0.3$}\\
\bottomrule
\end{tabular}
\end{table*}

\textbf{The observation gate carries no information.} Its safe settings accelerate too few calls
to matter, every setting worth deploying certifies a collapse, and at a matched rate it does not
beat a coin; an exogenous signal that cannot beat a coin is not a monitor. Unskipped computation does not see it either: the policy's own value head ranks
collapse no better than chance, inside the band reported for reactive
detectors~\cite{gatenotcache}, and a finiteness check catches one collapse in $31$.

\textbf{Granting and withdrawing are not symmetric.} An \emph{admission} gate grants permission
to reuse; a \emph{veto-only} check withdraws it but never grants it. A false veto gives up acceleration,
a false admission risks the task, and a fooled veto leaves the call exactly where an
unmonitored loop would have left it --- unsafe, but no less safe. \S\ref{sec:method} therefore
grants nothing.

\textbf{Neither axis suffices alone, and the oracle shows why.} Given every full forward and
the exact local action error, thresholding still loses twenty points, and the natural reading,
that local information is insufficient, is wrong: that oracle is an \emph{admission gate}
that re-arms after every rejection, so it keeps admitting chunks and small errors accumulate.
Hold the statistic fixed, make the response absorbing, and the same oracle costs one point.

That swing is real and it is not free. A rule that never forgets retires at its first
rejection, so what absorbing costs is set by when that rejection lands --- a property of how a
statistic's rejections \emph{cluster}, not of its rate. The rate-matched coin rejects
independently, so it admits under one call in expectation and the episode runs $\pi$: safety
under absorbing is never by itself evidence that a statistic carries information. The oracle's
rejections come in runs, so it still accelerates $2.0\%$ of the episode against
$15.3\%$ re-arming, where an independent rule at its rate would not keep even one
accelerated call in expectation --- and that is what its one point costs. The observation gate
at its safe settings, admitting early and rarely after, is why the column is not monotone in
the accelerated share. The same clustering is why the control is a coin matched to the gate's \emph{retirement}
rather than to its rate; on it the gate at its useful settings is indistinguishable from the
coin. What decides whether stopping is affordable is the statistic, which is the
argument for the last block: its check vetoes at most $1.0\%$ of the calls it examines
on the arms we ship, and far fewer on three of the four, so absorbing costs it almost nothing.
The two axes multiply, and the gating literature has varied only the first.

\textbf{What does work, and why it is not free.} Deviation from the policy the accelerator
replaced separates the two
populations by more than forty at the median --- $0.018$ against $0.81$ --- and ranks
catastrophe at $\mathrm{AUROC}\ 0.984$ over the $114$ configurations, $1.000$ inside each family
that has a collapse, their tails overlapping between $0.066$ and $0.332$ (Fig.~\ref{fig:frontier}a). Two coordinates sort Table~\ref{tab:map}: what the statistic measures, and what the
loop does when it fires. A quantity the accelerator itself sets, or one insensitive to what it
broke, fails however the loop responds. Rule direction is the obvious third candidate and the
table rules it out: the same statistic read as a veto-only check lands marginally \emph{worse}
than as a gate, because flipping the rule changes only the default, which now accelerates the
calls that had nothing cached to read. Direction decides the cost of being wrong, not whether
the monitor works, which is why \S\ref{sec:method} takes the veto-only side.

Table~\ref{tab:map}'s last block reads the statistic off the first row's cache, which decodes
from stale features, so Eq.~\eqref{eq:fnrmix} scores it as recompute-type: of the $160$
accelerated calls an episode offers, the cache damages $94$, $\delta^{\mathrm{free}}$
misses $1.9$ of those, and $\lambda$ misses all of them. The two misses are different
populations: what $\delta^{\mathrm{free}}$ lets through sits below $\theta$ by construction and
costs about a third of a point against a tolerance of $38$ chunks at the power-law
concavity \S\ref{sec:frontier} measures on Cosmos, while what $\lambda$ lets through is every chunk the cache
makes. Under \emph{absorbing} a vetoing check retires the
accelerator at its first veto, inside two accelerated calls here; the ladder retires a rung
instead.

\section{The Unexpired Plan}\label{sec:method}
The map says the informative statistic is a comparison against the replaced policy, and that
no monitor holds it for free. Both halves of this section attack the second clause.

\subsection{Setting and statistic}
At control call $t$ the policy $\pi$ maps observation $o_t$ to a chunk
$A_t=\pi(o_t)\in\mathbb{R}^{H\times d}$, of which the first $H_{\mathrm{exec}}<H$ actions
execute before the next call; write $m=\lfloor H/H_{\mathrm{exec}}\rfloor$. Every family
executes $H_{\mathrm{exec}}{=}2$ actions per call, so $m$ is set by the planning horizon
alone --- $2$, $5$, $2$ and $4$ on the four families, in Table~\ref{tab:main}'s column heads
--- and an episode of $L$ control calls costs $2L$ environment steps; a longer stride would
shorten $m$, and with it both the free coverage and Eq.~\eqref{eq:cost}'s ceiling. An accelerator is a configuration $c$ giving $\pi_c$ at per-call cost $C_c$ against
reference cost $C_f$, with $S=C_f/C_c$, and the controller runs at period $T\ge C_f$. Every
decision uses one scalar,
\begin{equation}
\delta\bigl(A,A^{\mathrm{ref}}\bigr)=
\frac{\lVert A-A^{\mathrm{ref}}\rVert_1}{\lVert A^{\mathrm{ref}}\rVert_1+\varepsilon},
\label{eq:dev}
\end{equation}
evaluated in the policy's normalised action space; the scale at which a safe configuration
parts from a collapse still differs by family (Fig.~\ref{fig:frontier}a), which
\S\ref{sec:exp} returns to. Two evaluations of the stochastic $\pi$ differ by a floor
$\delta_{\mathrm{self}}$, $0.011$ here.

\subsection{The free monitor: the unexpired plan}\label{sec:free}
A reference evaluation at control call $t$ leaves $H-H_{\mathrm{exec}}$ actions unused.
For the next $m-1$ calls, predictions of the actions they will execute are already in hand:
consecutive horizons overlap, and the plan has not expired. Write
$A^{\mathrm{ref}}_{t\to t+j}$ for stored rows $jH_{\mathrm{exec}}:(j+1)H_{\mathrm{exec}}$,
using zero-based, end-exclusive row slices. For $1\le j\le m-1$, the executed-prefix check
\begin{equation}
\delta^{\mathrm{free}}_{t+j}=\delta\bigl(\pi_c(o_{t+j})_{:H_{\mathrm{exec}}},\,A^{\mathrm{ref}}_{t\to t+j}\bigr),
\label{eq:free}
\end{equation}
costs no forward pass and reads the accelerator against the replaced policy's own output --- a
quantity the accelerator cannot set, which is what Table~\ref{tab:map}'s first block lacks. Two
coverages must be kept apart, both counting identities: \emph{free coverage}
$\phi=(m-1)/m$ is the share of calls a free check reaches, and since the remaining $1/m$ runs
the reference, which executes what it has just computed, the \emph{guarded} share is
$\phi+1/m=1$. Running the reference every $k>m$ calls leaves accelerated calls unreached and
the \emph{checked share} of them falls to $(m-1)/(k-1)$; Table~\ref{tab:main} reports the
guarded share, and Fig.~\ref{fig:frontier} prices every $k>m$.

Eq.~\eqref{eq:free} differs from a paid comparison. Before reference-norm normalisation,
write $e$ for accelerator error and $d$ for reference replanning drift: the free numerator is
$\lVert e+d\rVert_1$, the paid one $\lVert e\rVert_1$. Their difference is at most
$\lVert d\rVert_1$ in magnitude; cancellation can cause \emph{under}-reporting.
An accelerator reusing a
previously computed chunk emits the very plan
$\delta^{\mathrm{free}}$ compares against, so $e\approx-d$ and the monitor reads near zero
exactly when the policy is stalest (Fig.~\ref{fig:pipeline}b). The unexpired plan is structurally blind to lag. On the recompute branch the bias runs the other way and grows with the age of the
plan: on $\pi_{0.5}$, which reads it at four ages, the mean free reading rises from
$0.021$ at $j{=}1$ to $0.038$ at $j{=}4$, and the veto rate with it.

\textbf{The second free check.} Reuse, and the blends between, need a second check --- also
free and needing no reference --- posed as a \emph{degree} rather than a test of equality:
$\lambda_t=\min_{1\le s\le K}\delta(\pi_c(o_t),A_{t-s})$, the distance to the nearest
recently emitted chunk over a window of $K=8$ calls, which covers every plan still live
here. It reads the accelerator's own output, which
\S\ref{sec:map} disqualifies for admission --- but $\lambda$ only withdraws acceleration;
closed-loop evaluation must still price the stale fallback. A policy holding still emits near-identical consecutive chunks legitimately,
and so does the reference, so $\theta_\lambda$ sits below the lower tail of the reference's own
$\lambda$ under reference-against-reference replay: a false veto in under $1.4\%$ of
\emph{episodes}. The drift $\lVert d\rVert$ is calibrated from that replay and subtracted on the
recompute branch only, where the bias is upward; on the reuse branch $\delta^{\mathrm{free}}$ is
already blind, and that branch belongs to $\lambda$. A third mechanism escapes both: a bias
too small to be a repeat and too small on any call to cross $\theta$. One crossing is
enough, since no demotion is undone; \S\ref{sec:lim} prices what never crosses.

\subsection{The uncovered call}\label{sec:cost}
One call in $m$ has no unexpired plan, and there we \emph{run the reference and execute it}.
A veto costs no forward pass
either: the vetoed call executes the unexpired reference slice it was just compared against.
That slice was computed at an older observation, so the fallback is free in compute and not in
fidelity. Table~\ref{tab:main}'s open-loop row makes the same substitution at every call;
a veto substitutes selectively where the check fires.
In this policy-forward cost model, peak per-call cost is $C_f$, covered by $T\ge C_f$;
an actual controller must also budget comparison and control overhead.
Over $m$ calls the policy spends $(m-1)C_c+C_f$ against $mC_f$, so
\begin{equation}
S_{\mathrm{eff}}=\frac{m\,C_f}{(m-1)\,C_c+C_f}=\frac{m}{(m-1)/S+1}\;\le\;m ,
\label{eq:cost}
\end{equation}
with equality only when the accelerated calls cost nothing, a corner reached by stopping
computation between reference calls and executing the unexpired plan itself: the veto branch
applied unconditionally, which Table~\ref{tab:main} runs. A paid comparison at every checked
call spends $(m-1)(C_c+C_f)+C_f$ instead, and the same table prices that. Monitoring is therefore not free in compute: one reference
evaluation per $m$ calls is charged in every number we report, and the $m$ that supplies the
coverage caps the speedup.

\subsection{Selection and recovery}\label{sec:sel}
We score a candidate by the $95$th percentile of Eq.~\eqref{eq:dev} over its calls --- the tail,
since an accelerator exact on most calls and badly wrong on a few has a small mean --- and
deploy the cheapest whose score clears $\theta$. Scoring on a stream recorded from the
unaccelerated policy is biased, a stateful accelerator replayed on reference observations never
feeding its own error back in, so we roll each candidate out in closed loop under its own
actions and score it against the reference evaluated on the states it visits, iterating until
the score stops moving: deviations only, never rollout success, from the candidate's own state
distribution. Candidates are ordered by cost, so a veto steps to the next \emph{more
conservative} rung with both checks still active; the demotion is never undone within the
episode, and only exhausting the ladder returns the policy to $\pi$, on $1.8\%$ of
episodes here.

\subsection{Exposure}\label{sec:exposure}
\begin{figure*}[!t]\centering
\includegraphics[width=\textwidth]{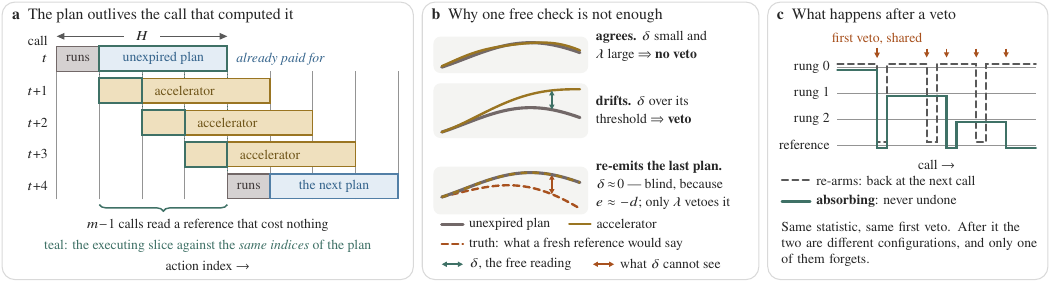}
\caption{\textbf{The method.} \textbf{(a)}~A reference plan spans $H$ indices while only
$H_{\mathrm{exec}}$ execute, so the next $m-1$ calls land on indices it has predicted and are read
against them at no forward pass. \textbf{(b)}~$\delta$ is Eq.~\eqref{eq:free}, $\lambda$ the repetition check of
\S\ref{sec:free}; re-emitting the plan it is compared against reads $\delta\approx0$, because
$e\approx-d$. \textbf{(c)}~A veto costs a rung before it costs the accelerator; a rung is a slower
configuration.}
\label{fig:pipeline}
\end{figure*}

Fig.~\ref{fig:pipeline} draws the coverage. Both checks run before execution on every covered
call, so at full coverage no accelerated chunk executes unexamined; detection is immediate, the
\emph{total} is not. Over an episode of $L$ calls, $200$ here, the expected number of
damaged chunks that execute is at most $L(1-1/m)\mathrm{FNR}(\theta)$, and safety requires that
\emph{expectation} to stay below a tolerance $n_c$; \S\ref{sec:frontier} shows when the tail
matters instead. A \emph{damaged} chunk is one the monitor should have vetoed and did
not, so we inject what such a chunk contains: the accelerator's output on a call whose deviation
exceeds $\theta$, replayed in place of the reference chunk --- the mildest chunk the monitor can
be wrong about, which makes $n_c$ an upper bound. $n_c$ is the largest number of those placeable
at random into an otherwise unaccelerated episode before paired success leaves the margin, by
bisection over $200$ paired episodes and reported as the \emph{minimum} over ten
seeds. The expression counts every
accelerated call, so demotion, which creates none, is already priced. Exposure grows with $m$, so the smallest tolerance is not the worst case
--- $\pi_{0.5}$, not PushT, is where the expectation comes closest.

\section{Experiments}\label{sec:exp}
\textbf{Protocol.} The four families of \S\ref{sec:map}, $1500$ paired episodes per
cell ($500$ on each of three seeds), $95\%$ paired bootstrap intervals clustered on initial state,
and a pre-stated $\pm2\pp$ equivalence margin. Every raw $S$ is a per-call compute
ratio $C_f/C_c$ at matched hardware and batch, not a FLOP count, and every $S_{\mathrm{eff}}$
composes those through Eq.~\eqref{eq:cost}; both are mean compute per call, not episode
duration (\S\ref{sec:lim} says what that buys). All are quoted against a baseline with unread future-frame decoding
already deleted, so no arm is credited for removing dead computation. We set $\theta=0.15$ at
the survivors' upper tail --- their 95th percentile is $0.16$ --- and well clear of the two
floors it must also pass, the policy's own seed-to-seed deviation and the $0.014$ offset the unexpired plan introduces. Fig.~\ref{fig:frontier}a shows what a single threshold costs: the populations part at a
different scale in each family, and on PushT $\theta$ admits two of five collapses that a
threshold set from PushT's own survivors would catch --- but that threshold needs PushT's
outcomes, which a deployment does not have. Rescuing the selected candidate without those
outcomes is the monitor's value on PushT; on the fifth family none of its outcomes had been seen. Every verdict in Tables~\ref{tab:map} and~\ref{tab:main} is unchanged for
$\theta\in[0.09,0.31]$, and holding any one family out of the tail that sets $\theta$
still lands inside that band, so no verdict turns on which families were seen. That is the
weaker half of what a held-out family tests; the stronger half, that the ladder, the candidate
set and the drift correction were designed with all four in view, is what the fifth family is
for. Table~\ref{tab:ablation} ran at the deployed threshold only.

\begin{table*}[t]\centering\papertable\small
\caption{Closed-loop comparison. Column heads: family, $m$, base success rate; first
row: paired half-width. \emph{guarded} is \S\ref{sec:free}'s guarded share; rows charging no reference call report raw $S$; \grey{grey} means unranked. $^{\dagger}$ marks
$|\Delta\mathrm{SR}|+h<2$ \emph{strictly}; \best{bold} and \second{underline} rank
$S_{\mathrm{eff}}$ among fully guarded arms whose point estimate is inside the margin on every
family.
The $\pi_{0.5}$ pair ($97.0$, $-91.0$) is the collapse of Fig.~\ref{fig:teaser}.}
\label{tab:main}
\setlength{\tabcolsep}{3.6pt}
\begin{tabular}{@{}lccccccccc@{}}
\toprule
& & \multicolumn{2}{c}{Cosmos $m{=}2$, $98.2$} & \multicolumn{2}{c}{$\pi_{0.5}$ $m{=}5$, $97.0$}
& \multicolumn{2}{c}{PushT $m{=}2$, $58.0$} & \multicolumn{2}{c}{MetaWorld $m{=}4$, $94.2$}\\
\cmidrule(lr){3-4}\cmidrule(lr){5-6}\cmidrule(lr){7-8}\cmidrule(lr){9-10}
Method & guarded & $\Delta$SR & $S_{\mathrm{eff}}$ & $\Delta$SR & $S_{\mathrm{eff}}$ & $\Delta$SR & $S_{\mathrm{eff}}$ & $\Delta$SR & $S_{\mathrm{eff}}$\\
\midrule
\emph{paired half-width} & & \multicolumn{2}{c}{$\pm1.1\pp$} & \multicolumn{2}{c}{$\pm0.9\pp$} & \multicolumn{2}{c}{$\pm2.6\pp$} & \multicolumn{2}{c}{$\pm1.4\pp$}\\[2pt]
\multicolumn{10}{@{}l}{\emph{unmonitored --- no guarantee at any operating point}}\\
Feature cache~\cite{worldcache_eccv}, selected & $0\%$ & $+0.1$$^\dagger$ & \grey{$1.40\times$} & $-0.3$$^\dagger$ & \grey{$1.20\times$} & $-0.9$ & \grey{$1.70\times$} & $-0.2$$^\dagger$ & \grey{$1.50\times$}\\
Chunk retrieval~\cite{actioncache}, selected & $0\%$ & $-0.6$$^\dagger$ & \grey{$1.70\times$} & $-1.4$ & \grey{$1.60\times$} & $-1.8$ & \grey{$2.40\times$} & $-0.9$ & \grey{$1.80\times$}\\
Learned tolerance~\cite{actionjnd} & $0\%$ & $-0.8$$^\dagger$ & \grey{$2.10\times$} & $-1.1$ & \grey{$1.90\times$} & $-3.1$ & \grey{$5.20\times$} & $-0.6$ & \grey{$3.40\times$}\\
Uniform step reduction, selected & $0\%$ & $-0.1$$^\dagger$ & \grey{$1.90\times$} & $-0.3$$^\dagger$ & \grey{$1.90\times$} & $-14.8$ & \grey{$47.0\times$} & $+0.0$$^\dagger$ & \grey{$10.0\times$}\\
Uniform step reduction, matched to our speed & $0\%$ & $+0.0$$^\dagger$ & \grey{$1.55\times$} & $-0.2$$^\dagger$ & \grey{$1.67\times$} & $-0.4$ & \grey{$1.75\times$} & $+0.1$$^\dagger$ & \grey{$3.09\times$}\\
Cross-call cache, own gate & $0\%$ & $-87.3$ & \grey{$1.60\times$} & $-91.0$ & \grey{$1.70\times$} & $-58.0$ & \grey{$4.00\times$} & $-44.1$ & \grey{$2.10\times$}\\
\midrule
\multicolumn{10}{@{}l}{\emph{fully monitored --- every call checked}}\\
\emph{Reference every call} ($\pi$ re-run against itself) & $100\%$ & $-0.3$$^\dagger$ & \second{$1.00\times$} & $-0.4$$^\dagger$ & \second{$1.00\times$} & $-0.7$ & \second{$1.00\times$} & $-0.3$$^\dagger$ & \second{$1.00\times$}\\
Paid deviation at every call (cf.~\cite{stac}) & $100\%$ & $-0.1$$^\dagger$ & $0.87\times$ & $-0.2$$^\dagger$ & $0.71\times$ & $+0.9$ & $0.97\times$ & $+0.0$$^\dagger$ & $0.93\times$\\
Chunk repetition alone ($\lambda$) & $100\%$ & $-0.6$$^\dagger$ & \grey{$3.40\times$} & $-0.7$$^\dagger$ & \grey{$2.00\times$} & $-9.4$ & \grey{$18.6\times$} & $-0.4$$^\dagger$ & \grey{$10.2\times$}\\
\midrule
\multicolumn{10}{@{}l}{\emph{the schedule without us --- what the reference period alone is worth}}\\
Open-loop replay, no accelerator ($C_c{=}0$) & --- & $-1.9$ & \grey{$2.00\times$} & $-12.4$ & \grey{$5.00\times$} & $-9.8$ & \grey{$2.00\times$} & $-4.1$ & \grey{$4.00\times$}\\
Same schedule, monitor off & $1/m$ & $-0.4$$^\dagger$ & \grey{$1.55\times$} & $-0.6$$^\dagger$ & \grey{$1.67\times$} & $-6.2$ & \grey{$1.90\times$} & $-0.3$$^\dagger$ & \grey{$3.09\times$}\\
\midrule
\best{Unexpired plan (ours)} & $100\%$ & \best{$+0.1$}$^\dagger$ & \best{$1.55\times$} & \best{$-0.2$}$^\dagger$ & \best{$1.67\times$} & \best{$+1.2$} & \best{$1.75\times$} & \best{$+0.1$}$^\dagger$ & \best{$3.09\times$}\\
\bottomrule
\end{tabular}
\end{table*}

\textbf{What are the right baselines?} Without the plan, a reference-reading monitor must stop accelerating and be the reference,
or refresh on its own schedule and pay for both on the calls it refreshes, as the
nearest prior monitor does~\cite{gatenotcache}; under Eq.~\eqref{eq:cost}'s cost model that
is the paid-deviation row at its refresh period, so we did not re-run it; its wall-clock claim
is about latency, which \S\ref{sec:lim} says we do not buy. Substituting the free reading for the paid one is this method's single
approximation, and the \emph{paid deviation} row prices it: run the reference at every checked
call and compare the accelerator against it, and the verdict lands within a half-width of ours
on all four families while the arm runs below the unaccelerated policy. The statistic is not
what is new here. Its price is. $\lambda$ alone keeps the speed and loses the sight: it is the
fastest fully guarded arm, certifies exactly where everything else certifies, and gives up
$9.4\pp$ on PushT, because Eq.~\eqref{eq:fnrmix} makes it blind to the
mechanism our selection lands on.

The last of them needs no monitor and no accelerator. If a vetoed call can execute the unexpired
plan, the whole period can: run the reference every $m$ calls and play out what it predicted.
That is Eq.~\eqref{eq:cost}'s ceiling rather than a point below it, so it is faster than us
everywhere, and what it gives up is the observation. None of the four certifies it, and the
loss grows the longer the loop stays open --- except on PushT, where one unobserved call in
two costs nearly what four in five cost on $\pi_{0.5}$; four families are four confounded
points, so what a plan costs to leave running is not set by $m$ alone. The line from the
reference at $1.00\times$ to open loop at $m$ is what the reference period alone offers at any
speed between, with no accelerator and no monitor; at our own speed we sit above it on every
family, furthest on PushT, and so, on every family, does an unmonitored step reduction matched
to our speed. The unmonitored rows check nothing
at all; uniform step reduction
is the hardest of them to fool, making no per-call decision, yet it loses $14.8\pp$ on PushT at its selected setting, and nothing
at deployment time separates that case from the three where it ties us. Matched to our speed
instead, it holds PushT's point estimate as well, and paired against our row on identical
initial states it trails by $1.6\pp$ against a between-arm half-width of
$1.2\pp$: what separates our row there is not speed a hindsight choice could not have
bought, but a margin no hindsight was needed for. On the fifth family the same control loses
outright (Table~\ref{tab:heldout}).

\textbf{Is it the monitor, or just the schedule?} The matched-schedule row answers it. With
every check off it matches our $S_{\mathrm{eff}}$ to the printed decimals on the three families
where vetoes are rare; its speed is Eq.~\eqref{eq:cost} at Fig.~\ref{fig:frontier}'s raw $S$; the \emph{chunk
repetition alone} row runs the same candidate and, $\lambda$ never firing on a recompute-type
mechanism, is that candidate unmonitored. On PushT our row and the matched
schedule differ by $7.4\pp$ on identical initial states,
against a paired half-width of $1.2\pp$ between them: halving an arm's exposure is the
right direction and nowhere near enough.

\textbf{What the daggers do and do not say.} A dagger means
only that a paired interval fits inside the margin, which a hand-picked accelerator usually
manages; the \emph{guarded} column separates our row. PushT certifies nowhere, ours included,
its half-width being wider than the margin; a one-sided reading would pass us, but on the same
column it fails \emph{$\pi$ re-run against itself}, whose interval reaches $-3.3$, and a test
the reference cannot pass measures the episode count, not the arm. We report PushT undecided. On the other three families the matched-schedule row is already inside the margin, so
what Table~\ref{tab:main} establishes about the monitor rests on PushT and on the two tests
that follow, not on those columns.

\textbf{Does it transfer off the configurations we chose?} We applied the monitor unchanged to
four arms the selection rejected, one per family (Table~\ref{tab:transfer}). None helped set
$\theta$, and none survived it; the reference period
alone, with every check off, leaves most of each loss in place. Monitored, three certify
and the PushT arm does not, because PushT certifies nothing; the cost lands on speed, the ladder
being exhausted on $9.6\%$ of transfer episodes against $1.8\%$ on ours. The
Cosmos arm is the one reuse-type mechanism of the four and the only place $\lambda$ does any
work: without it the monitor keeps only what the schedule keeps, because $\delta^{\mathrm{free}}$ reads a re-emitted chunk at zero
--- the closed-loop case Table~\ref{tab:ablation} cannot show, every arm we ship being
recompute-type.

\begin{table}[!tb]\centering\papertable\fontsize{8.5}{10}\selectfont
\caption{Four configurations the selection rejected, one per family, monitored unchanged. $\Delta$SR against each family's base, half-widths as in Table~\ref{tab:main};
\emph{schedule} runs the arm on the reference period with every check off; \emph{no
$\lambda$} keeps only Eq.~\eqref{eq:free}; the $\pi_{0.5}$ arm is Fig.~\ref{fig:teaser}'s
collapse. The $-91.0$ is measured.}
\label{tab:transfer}
\setlength{\tabcolsep}{1.9pt}
\begin{tabular}{@{}lcccccc@{}}
\toprule
& & \multicolumn{3}{c}{$\Delta$SR, monitor:} & \multicolumn{2}{c}{\best{monitored}}\\
\cmidrule(lr){3-5}\cmidrule(lr){6-7}
Rejected arm (family) & raw $S$ & none & schedule & no $\lambda$ & $\Delta$SR & $S_{\mathrm{eff}}$\\
\midrule
Cross-call cache ($\pi_{0.5}$) & \grey{$1.70\times$} & $-91.0$ & $-79.2$ & $-0.3$$^\dagger$ & \best{$-0.3$}$^\dagger$ & \best{$1.05\times$}\\
Step reduction (PushT) & \grey{$88.9\times$} & $-58.0$ & $-37.7$ & $-1.4$ & \best{$-1.4$} & \best{$1.30\times$}\\
Token reuse (Cosmos) & \grey{$2.60\times$} & $-18.4$ & $-12.0$ & $-12.0$ & \best{$-0.6$}$^\dagger$ & \best{$1.10\times$}\\
Feature cache (MetaWorld) & \grey{$2.20\times$} & $-9.0$ & $-7.5$ & $-0.3$$^\dagger$ & \best{$-0.3$}$^\dagger$ & \best{$1.18\times$}\\
\bottomrule
\end{tabular}
\end{table}

\textbf{A fifth family, after the design was frozen.} Table~\ref{tab:main}'s own verdict on the
monitor rests on its one undecidable column, and everything above was designed with four
families in view. So we froze $\theta$, $\theta_\lambda$, $\lambda$'s window, the ladder, the
drift correction and the candidate scoring, and ran a family none of them had seen: the same
Diffusion Policy on ManiSkill3's tight-tolerance peg insertion~\cite{maniskill3}, at $H{=}6$
and $H_{\mathrm{exec}}{=}2$, so $m{=}3$. We chose it where the schedule was expected
to fail: it is contact-rich, so a plan expires fast. And its simulator is GPU-parallel, so
paired episodes cost little enough to run four times as many as elsewhere, which at PushT's
clustering is what brings the half-width inside the margin.

Table~\ref{tab:heldout} is what came back. Selection lands on a feature cache, and unmonitored
it loses $8.7\pp$; a step reduction matched to our speed loses less and is still
decidably outside the margin, playing the old plan out open-loop loses most of all. The matched
schedule recovers only part of the cache's loss and stays well outside, and the monitor, on
the same schedule, brings the family inside the margin with the dagger, at a cost of six percent of the
schedule's speed; the paid comparison agrees, below the reference's speed. This is the one column in the paper where the monitor is
necessary at its speed, decidable and correct at once, and nothing in the method was tuned to it.

\begin{table}[!tb]\centering\papertable\small
\caption{A fifth family, run after the design was frozen: the Diffusion
Policy~\cite{diffusionpolicy} on ManiSkill3's tight-tolerance peg insertion~\cite{maniskill3}.
Columns are Table~\ref{tab:main}'s, and the first body row gives the paired half-width, here
inside the margin.}
\label{tab:heldout}
\setlength{\tabcolsep}{5pt}
\begin{tabular}{@{}lccc@{}}
\toprule
Method & guarded & $\Delta$SR & $S_{\mathrm{eff}}$\\
\midrule
\multicolumn{4}{@{}l}{\emph{ManiSkill3 insertion, $m{=}3$, base $71.3$, $6000$ episodes}}\\
\emph{paired half-width} & & $\pm1.2\pp$ & \\[2pt]
Feature cache, selected & $0\%$ & $-8.7$ & \grey{$4.60\times$}\\
Step reduction, matched to our speed & $0\%$ & $-3.9$ & \grey{$1.96\times$}\\
\midrule
\emph{Reference every call} & $100\%$ & $-0.3$$^\dagger$ & \second{$1.00\times$}\\
Paid deviation at every call & $100\%$ & $-0.1$$^\dagger$ & $0.87\times$\\
Chunk repetition alone ($\lambda$) & $100\%$ & $-8.7$ & \grey{$4.60\times$}\\
\midrule
Open-loop replay ($C_c{=}0$) & --- & $-11.5$ & \grey{$3.00\times$}\\
Same schedule, monitor off & $1/m$ & $-6.7$ & \grey{$2.09\times$}\\
\midrule
\best{Unexpired plan (ours)} & $100\%$ & \best{$-0.2$}$^\dagger$ & \best{$1.96\times$}\\
\bottomrule
\end{tabular}
\end{table}

\textbf{Ablation.} Table~\ref{tab:ablation} crosses the two halves, and the crossed cell reads
down the column rather than across: switching the monitor on rescues off-policy selection by
more than any half-width in this paper, so the monitor and not the selection is what removes
the collapse. That does not make the rescued arm \emph{safe} --- PushT certifies nothing, ours
included --- but it does order them: on identical initial states the two selections differ
by $3.1\pp$ against a paired half-width of $1.2\pp$, so on-policy is the safer as
well as the faster. That one veto
an episode is also the whole $0.15\times$ between the matched schedule and the row we
deploy, and latching it to the reference instead of to a rung costs PushT another
$0.29\times$, the other three unchanged. Of the three rows at $1.93\times$, one is the deployed row and one repeats it
(\emph{no}~$\lambda$, below); the third, \emph{mean instead of tail}, reaches the same speed the
opposite way, a fast candidate paying for itself in demotions where ours, slower, rarely demotes. Re-arming instead of
absorbing keeps the schedule's speed and gives back $2.8\pp$ on PushT, the rung
re-admitted after every veto: on our statistic the response matters only where vetoes are not rare.

\emph{Mean instead of tail} is also the one row the monitor does not rescue, and missed calls
cannot account for it: at the composite
rate its accelerated calls expose under a seventh of the damaged chunks the tolerance absorbs,
even counting every one as damaged. The damage is in
the calls that never cross $\theta$: its median per-call deviation is $0.09$ against
$0.02$ on the shipped arm, three fifths of the way to the threshold, pushed there
by a bias no single call is large enough to expose. No per-call rule at our threshold reaches such a bias (\S\ref{sec:lim}); scoring on the tail
keeps such arms out for free. The \emph{no}~$\lambda$ row repeats the deployed row,
$\lambda$ earning its place only on the reuse-type transfer arm.

\begin{table}[!tb]\centering\papertable\small
\caption{Ablation: monitor crossed with selection above the rule, one design choice knocked
out below it. $\Delta$SR is PushT's, the family that separates them; $\bar S_{\mathrm{eff}}$ is
the geometric mean over all four, which Eq.~\eqref{eq:cost} does not commute with; \emph{no
monitor} charges no reference call, so it reports raw $S$. The \emph{no}~$\lambda$ row repeats the
deployed row exactly.}
\label{tab:ablation}
\setlength{\tabcolsep}{5pt}
\begin{tabular}{@{}lcccc@{}}
\toprule
& \multicolumn{2}{c}{no monitor} & \multicolumn{2}{c}{$+$ monitor}\\
\cmidrule(lr){2-3}\cmidrule(lr){4-5}
Selection & PT $\Delta$SR & $\bar S_{\mathrm{eff}}$ & PT $\Delta$SR & $\bar S_{\mathrm{eff}}$\\
\midrule
off-policy & $-31.2$ & \grey{$7.56\times$} & $-1.9$ & $1.31\times$\\
\best{on-policy} & $-9.4$ & \grey{$5.99\times$} & \best{$+1.2$} & \best{$1.93\times$}\\
\midrule
mean instead of tail & $-38.4$ & \grey{$9.12\times$} & $-22.8$ & $1.93\times$\\
no $\lambda$ & $-9.4$ & \grey{$5.99\times$} & $+1.2$ & $1.93\times$\\
latch instead of ladder & $-9.4$ & \grey{$5.99\times$} & $+1.2$ & $1.85\times$\\
re-arm instead of absorb & $-9.4$ & \grey{$5.99\times$} & $-1.6$ & $1.97\times$\\
\bottomrule
\end{tabular}
\end{table}

\subsection{Coverage buys safety where you cannot price it}\label{sec:frontier}
\begin{figure*}[!t]\centering
\includegraphics[width=\textwidth]{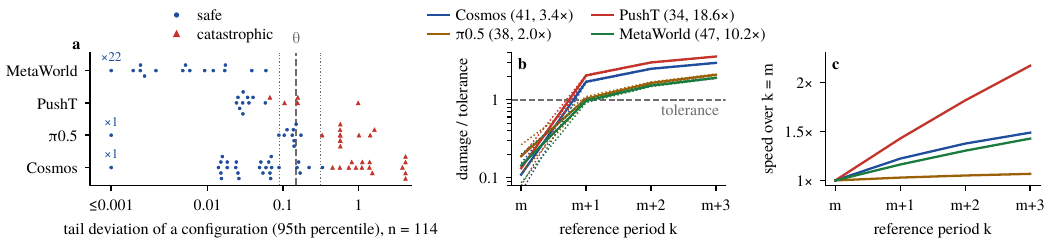}
\caption{\textbf{(a)}~The tail deviation (Eq.~\eqref{eq:dev}'s 95th percentile over an
episode) of each of the $114$ configurations of \S\ref{sec:map}, catastrophic when it loses
ten points or more; values below $0.001$ pool at the left edge with their count; values within a marker width overprint. The dashed
line is $\theta$, the dotted lines the range over which every verdict in Tables~\ref{tab:map}
and~\ref{tab:main} holds. \textbf{(b)}~Stepping the reference period $k$ past full coverage: the share of the
tolerance $n_c$ that Eq.~\eqref{eq:feas} spends at $p_d{=}1$, at the composite miss rate
(solid) and across its interval (dotted), for Cosmos, $\pi_{0.5}$, PushT and MetaWorld at
tolerances of $41$/$38$/$34$/$47$ chunks. \textbf{(c)}~What the
step buys, Eq.~\eqref{eq:cost} at raw $S$ of $3.4$/$2.0$/$18.6$/$10.2$
before the ladder's vetoes; PushT deploys at Table~\ref{tab:main}'s $1.75\times$.}
\label{fig:frontier}
\end{figure*}

Running the reference every $k\ge m$ calls checks $(m-1)/(k-1)$ of the accelerated calls, so
coverage looks like a dial trading against speed. With $p_d$ the rate at which an accelerated
call damages its chunk, the damage that executes over $L$ calls is
\begin{equation}
E(k)\;=\;p_d\,L\Bigl[\underbrace{\tfrac{k-m}{k}}_{\text{uncovered}}
\;+\;\underbrace{\tfrac{m-1}{k}\,\mathrm{FNR}(\theta)}_{\text{missed}}\Bigr],
\label{eq:feas}
\end{equation}
where the first term carries no $\mathrm{FNR}$ because an uncovered call is never examined. We
do not know $p_d$ --- \S\ref{sec:map} is the finding that nothing before deployment does ---
but it multiplies both terms, so it cancels from the \emph{ratio} between periods:
\begin{equation}
\frac{E(m{+}1)}{E(m)}\;=\;\frac{m\bigl[1+(m-1)\mathrm{FNR}\bigr]}{(m+1)(m-1)\,\mathrm{FNR}} .
\label{eq:thresh}
\end{equation}
Across the four families, by Eq.~\eqref{eq:thresh}, one step off full coverage multiplies
expected damage by $5.4$ to $15.5$ at the composite rate, and by at least four anywhere in its interval. That
first step buys at most $1.43\times$ in speed; the steps after it buy more and spend
more (Fig.~\ref{fig:frontier}b,c).

\textbf{Where the composite miss rate comes from.} A chunk escapes only if both checks miss,
and the two are mirror images, so the joint rate composes per mechanism:
\begin{equation}
\mathrm{FNR}(\theta)=\sum\nolimits_j w_j\,
\mathrm{FNR}^{\delta}_j\,\mathrm{FNR}^{\lambda}_j=0.045,
\label{eq:fnrmix}
\end{equation}
over pure repeat, blend and recompute at weights $25/12/63\%$, with marginals
$(1.00,0.06)$, $(0.50,0.29)$ and $(0.02,1.00)$ --- $\delta^{\mathrm{free}}$
blind to a repeat, $\lambda$ to a recompute. Each $w_j$ is the share of the sweep's accelerated
calls whose mechanism is $j$, and each marginal is one check's miss rate on injected chunks of
that mechanism, over the same $200$ paired episodes and ten seeds as $n_c$. The
product assumes independence, which we state
rather than test; over every dependence structure with those marginals the composite stays in
$[0.027,0.063]$ outward-rounded, two of the three mechanisms being pinned outright, so
the assumption rides on the blend alone.

\textbf{Turning the ratio into a verdict needs $p_d$, and we bound it.}
Fig.~\ref{fig:frontier}b prices every accelerated call at $p_d{=}1$, and the \emph{same
schedule, monitor off} arm shows how loose that is: it runs its whole accelerated population
unexamined, would leave the margin everywhere at that rate, and stays inside on three families,
which puts $p_d$ near $0.04$--$0.09$ there if loss were linear in damage. It is
not: \S\ref{sec:exposure}'s bisection finds that half the injected damage already buys
$65\%$ of the loss on Cosmos, so loss is concave in damage and those three rates are
ceilings. 

Substituting that bound into Eq.~\eqref{eq:feas} cancels $n_c$: $k{=}m{+}1$ clears the
tolerance by $2(m^2{-}1)/\bigl(m\,\ell\,[1+(m{-}1)\mathrm{FNR}]\bigr)$, the $2$ being the
margin and $\ell$ the monitor-off loss in points, so neither the seed minimum nor the injected
severity enters it. On the three families where the monitor-off arm stays inside the margin, each $\ell$ is
smaller than its own half-width, so we quote the far end of its interval, and there $k{=}m{+}1$ still clears $n_c$ by at least $1.9\times$
while running up to $1.22\times$ faster than we ship. The trade is open exactly where
the monitor proved unnecessary. One family stays open even at $p_d{=}1$ (Fig.~\ref{fig:frontier}b): MetaWorld's fifth call
sits just under its tolerance at the composite rate and just over it at the top of that rate's
interval --- a warning, not a clearance. The figure prices at the sweep's composite rate, while
every arm we ship is recompute-type, where Eq.~\eqref{eq:fnrmix} gives $0.02$, so Eq.~\eqref{eq:thresh}'s ratio there is about twofold conservative for them. We ship $k{=}m$ for transfer rather than for
the bound: $p_d$ belongs to the arm selection landed on, \S\ref{sec:map} is the evidence that
nothing beforehand predicts it, and full coverage leaves $n_c$ roughly sevenfold above $E(m)$ in
geometric mean, against a rate the deployment cannot see.

\section{Limitations}\label{sec:lim}
Free coverage is a property of $m$: a policy that executes its whole plan gets none, the speedup is capped at the reference period, and the method pays best at large $m$.
A per-call compute ratio buys neither episode time (the period is fixed) nor peak forward cost ($T\ge C_f$ pins it at the reference): it may save energy or increase policy-serving capacity, but we measure neither. We deploy on the expectation because nothing we measure separates independent
misses from a sustained excursion: treat them as Poisson runs of $20$ calls at $p_d{=}1$, $0.36$ runs an episode
on $\pi_{0.5}$, and the deployed tail of the count rises to $0.05$.
Eq.~\eqref{eq:feas} prices \emph{per-call} misses, so a bias too small to reach $\theta$ on any call adds nothing to $\mathrm{FNR}$, and a rung is retired only by a
veto; such a bias is bounded by candidate scoring, not by any per-call rule. $p_d$ is calibrated on random damage while what escapes $\mathrm{FNR}$ is selected for being
undetectable, and every accelerated call is priced at one $p_d$ while the uncovered call is the
stalest of its period.
$\theta$ is read off the deviation percentiles of the $114$ configurations of \S\ref{sec:map};
the arms of \S\ref{sec:exp} come from the same four families but not from that sweep, so the threshold is fitted to the families, not the arms, and distribution-free risk control~\cite{ltt} would need more than one held-out family. All closed-loop evidence here is in simulation, not on the real-robot distributions these
policies serve~\cite{droid}.

\section{Conclusion}
Priced in closed-loop success rather than feature distance, the monitors a diffusion policy is offered buy less than their logs report: an accelerator's gate certifies its own
product, an exogenous gate does not beat random reuse, and exact local error at full compute
still loses twenty points under a re-arming rule. What works is a comparison against the replaced policy, under a response that does not forget --- and a chunked
policy is already holding one: its unexpired plan. Harvesting it costs no forward pass and caps speed at the reference period. The core monitor implementation and project materials are available through the linked project website.

\bibliographystyle{IEEEtran}
\bibliography{refs}
\end{document}